\documentclass[letterpaper, 10 pt, conference]{ieeeconf}

\usepackage[utf8]{inputenc} % allow utf-8 input
\usepackage[T1]{fontenc}    % use 8-bit T1 fonts
\usepackage{cite}
\makeatletter
\let\NAT@parse\undefined
\makeatother
\usepackage{hyperref}
\usepackage{url}            % simple URL typesetting
\usepackage{booktabs}       % professional-quality tables
\usepackage{amsfonts}       % blackboard math symbols
\usepackage{nicefrac}       % compact symbols for 1/2, etc.
\usepackage{microtype}      % microtypography
\usepackage{xcolor}         % colors
\usepackage{amsmath}
\usepackage{amssymb}
\usepackage{logicpuzzle}
\usepackage{stmaryrd}
\usepackage{algorithmic}
\usepackage{algorithm}
\usepackage{makecell}

\usepackage{amsmath,amsfonts}
\usepackage{algorithmic}
\usepackage{algorithm}
\usepackage{array}
\usepackage[caption=false, font=footnotesize]{subfig}
\usepackage{textcomp}
\usepackage{stfloats}
\usepackage{url}
\usepackage{verbatim}
\usepackage{graphicx}
\usepackage{epstopdf}
\usepackage[utf8]{inputenc}
\usepackage[T1]{fontenc}
\usepackage{booktabs} 
\usepackage{diagbox}  
\usepackage{multirow} 
\usepackage{threeparttable} 
\usepackage{epstopdf} 
\usepackage{colortbl}  % cellcolor
\usepackage{hhline}
\usepackage[inkscapelatex=false]{svg}
\definecolor{codegreen}{rgb}{0,0.5,0}
\definecolor{codeblue}{rgb}{0,0,0.9}
\definecolor{codeblues}{rgb}{0,0,0.4}
\definecolor{codegray2}{rgb}{0.4,0.4,0.4}
\definecolor{codegray}{rgb}{0.9,0.9,0.9}

\usepackage{color, xcolor}

\newcounter{ALC@tempcntr}% Temporary counter for storage
\newcommand{\LCOMMENT}[1]{%
    \setcounter{ALC@tempcntr}{\arabic{ALC@rem}}% Store old counter
    \setcounter{ALC@rem}{1}% To avoid printing line number
    \item //#1 % Display comment + does not increment list item counter
    \setcounter{ALC@rem}{\arabic{ALC@tempcntr}}% Restore old counter
}%

\usepackage{fontawesome}
\definecolor{Dred}{HTML}{C81D31}
\definecolor{Dgreen}{HTML}{588E31}
\definecolor{codegray}{rgb}{0.9,0.9,0.9}

\IEEEoverridecommandlockouts

\begin{document}
\title{CANDO: Conservative Action-Noise Directional Optimization for Residual Reinforcement Learning}

\author{Zeyu Gao$^{1,2*}$, Chengkai Hou$^{3*}$, Chaoyue Li$^{4}$, Zhuangzhe Wu$^{3,4}$, Jiaming Liu$^{3,4}$, 
\quad
\\
Yao Mu$^{5}$, Peng Jia$^{4}$, Li Zhang$^{1,2\dagger}$ and Shanghang Zhang$^{3\dagger}$
% \quad
\thanks{
Zeyu Gao* and Chengkai Hou* are co-first authors. Shanghang Zhang$^{\dagger}$ and Li Zhang$^{\dagger}$ are the corresponding authors.}
\thanks{
$^1,^2$ Zeyu Gao and Li Zhang are with the School of Data Science, Fudan University, Shanghai 200433, China, and also with Shanghai Innovation Institute, Shanghai 200233, China (e-mail: gaozeyu2026@sii.edu.cn; zhangfd@fudan.edu.cn)}
\thanks{
$^3$ Chengkai Hou, Zhuangzhe Wu, Jiaming Liu and Shanghang Zhang are with State Key Laboratory of Multimedia Information Processing, School of Computer Science, Peking University, Beijing 100871, China (e-mail: 2501111947@stu.pku.edu.cn; shanghang@pku.edu.cn)}
\thanks{
$^4$ Chaoyue Li, Zhuangzhe Wu, Jiaming Liu and Peng Jia are with Simplexity Robotics, Beijing 100080, China}
\thanks{
$^5$ Yao Mu is with the School of Computer Science, Shanghai Jiao Tong University, Shanghai 200240, China}
}
\maketitle

%===============================================================================

\begin{abstract}
Pretrained generative policies provide strong behavior priors for robot control, but adapting them efficiently with online reinforcement learning remains difficult. A central challenge is how to improve a frozen generative policy without destroying the structure learned from offline demonstrations. Noise-space residual methods address this by modifying the policy's input noise, but existing noise-space methods typically regulate adaptation through fixed latent feasible regions, which may exclude valuable noise edits. In contrast, action-space residual methods are more sample-efficient, but their efficiency can still be further improved. We propose Conservative Noise Directional Optimization (CNDO), a noise-space residual RL method that improves a frozen generative policy by learning state- and noise-conditioned directions around Gaussian latent samples. CNDO removes predefined latent feasible regions and instead regulates noise-space adaptation through the soft conservative constraint, which consists of conservative value estimation and gated execution. We further introduce Conservative Action-Noise Directional Optimization (CANDO), which adds an action-space residual, combining structured latent-space improvement with local action correction. Experiments show that, once CNDO narrows the gap between noise-space and action-space residual learning, CANDO further improves over single-space optimization, achieving 76.1 NAUC and a 91.4\% final success rate, surpassing the best single-space results by 6.6 NAUC and 4.1\% success rate. More demos and info are available on https://cando-rl.github.io/.

% More demos and code are available on \href{https://sites.google.com/view/dag-plan}{https://sites.google.com/view/dag-plan}.
\end{abstract}

%===============================================================================

\section{Introduction}
\label{sec:introduction}

Learning robot policies from offline demonstrations provides an effective way to acquire complex manipulation skills without manually designing task-specific controllers. In particular, generative policies based on diffusion models and conditional flow matching can represent complex and multi-modal action distributions and have demonstrated strong performance across a broad range of robotic tasks~\cite{chi2025diffusion,chisari2025pointflowmatch}. However, behavior cloning is inherently limited by the coverage of demonstration data and is susceptible to covariate shift during closed-loop deployment~\cite{ross2011reduction}. Online reinforcement learning (RL) provides a natural means to further improve pretrained policies through task feedback, but directly fine-tuning high-capacity generative policies typically requires optimization through the full generation process, which can introduce substantial computational cost and training instability~\cite{ICLR2025_c0749c39,wagenmaker2025steering}. This motivates lightweight adaptation methods that keep the pretrained policy frozen while learning only a task-specific adaptation module.

Residual reinforcement learning provides a natural framework for this setting by keeping a base policy fixed and learning only an additional residual correction~\cite{johannink2019residual}. For a frozen generative policy, residual adaptation can operate in two distinct spaces. In noise space, the input noise is modified before generation to alter the behavior produced by the frozen policy; in action space, the generated action is directly corrected after decoding. Diffusion Steering via Reinforcement Learning (DSRL) represents the former approach by treating the input noise as the action of a latent-space Markov decision process~\cite{wagenmaker2025steering}. In contrast, Distribution Contractive Reinforcement Learning (DICE) learns a lightweight residual over the generated action and combines residual optimization with value-guided action selection~\cite{sun2026from}. Although both spaces provide viable interfaces for online policy improvement, two important questions remain: can noise-space residual learning achieve sample efficiency comparable to strong action-space residual learning, and can the two spaces be jointly optimized to further improve policy learning?

Existing noise-space methods are substantially less sample-efficient than DICE on most of the tasks considered in our experiments. We find that this gap is partly associated with how noise-space policy updates are regulated. DSRL restricts latent actions to a bounded region, Latent Policy Steering (LPS) projects them onto the typical Gaussian hypersphere, and Lagrangian Perturbation Diffusion Steering (LP-DS) retains Gaussian noise samples while constraining their residual perturbations through a Lagrangian trust region~\cite{wagenmaker2025steering,im2026lps,simsir2026lagrangian}. Such mechanisms improve training stability by explicitly limiting the region that the noise policy can explore. However, latent displacement is only indirectly related to the quality of the behavior produced after decoding. Consequently, a predefined geometric region may reject large but valuable latent edits while still admitting small edits that do not improve the resulting behavior. This suggests that noise-space adaptation should be regulated according to the quality of the decoded behavior, rather than solely by a predefined geometric region in latent space.

\begin{figure*}[t]
    \centering
    \includegraphics[width=1.0\textwidth]{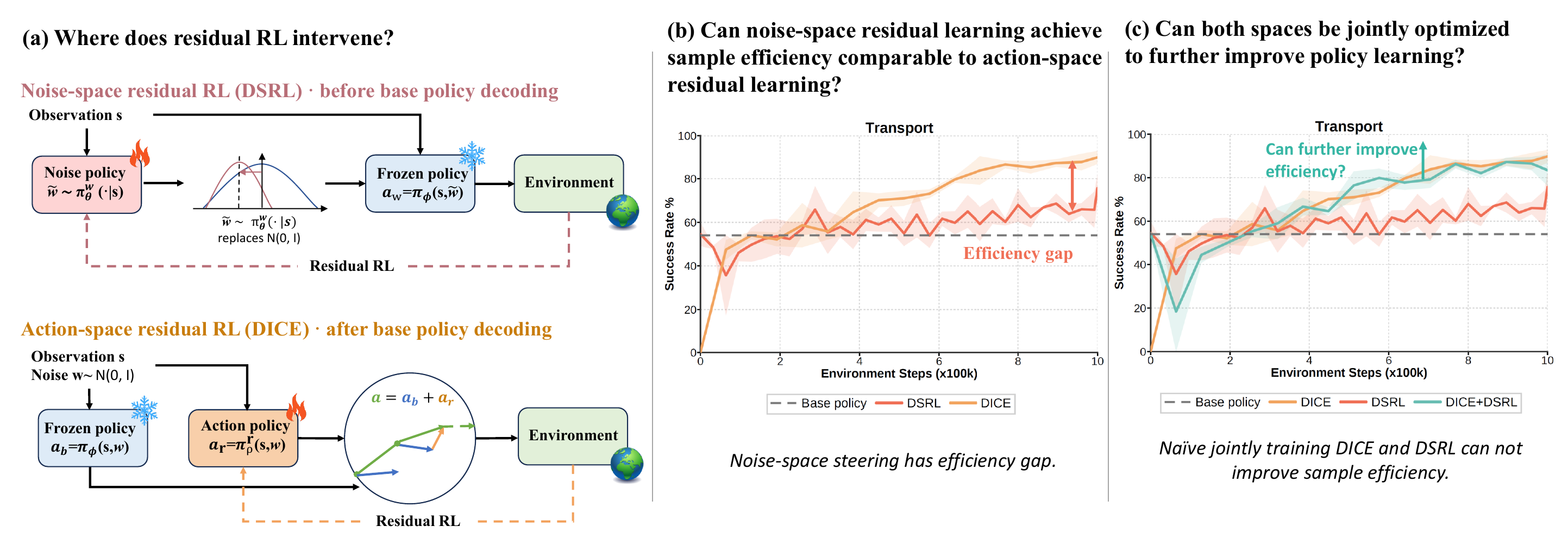}
    \vspace{-23pt}
    \caption{Motivation for residual learning in noise and action spaces.
    (a) Noise-space residual RL edits the input noise before frozen-policy decoding, while action-space residual RL corrects decoded actions.
    (b) Existing noise-space methods show a clear sample-efficiency gap relative to DICE.
    (c) Naively combining DSRL and DICE provides no additional improvement.}
    \vspace{-10pt}
    \label{fig:intro_overview}
\end{figure*}

To address this gap, we propose \textbf{Conservative Noise Directional Optimization (CNDO)}, which removes the predefined latent feasible region and instead evaluates and regulates noise edits through their decoded behaviors. Removing hard geometric restrictions enables broader noise-space exploration, while increasing the risk of unstable optimization. CNDO introduces a soft conservative constraint consisting of two components: a conservative critic and gated execution. The conservative critic suppresses unsupported value overestimation when optimizing noise directions, while the gate selectively executes proposed edits according to target-critic value estimates. In this way, CNDO expands the effective optimization range in noise space while preventing unrestricted exploration from destabilizing policy learning. 

% Our experiments show that CNDO substantially improves the sample efficiency of noise-space residual learning, outperforming existing noise-space methods and reaching performance comparable to the strong action-space residual method DICE.

We further combine CNDO with action-space residual learning and introduce \textbf{Conservative Action-Noise Directional Optimization (CANDO)}. The two residual branches affect the final policy through different pathways: the noise-space residual modifies the behavior generated through the frozen policy, whereas the action-space residual directly corrects the generated action. However, having access to both adaptation spaces does not necessarily guarantee additional gains from joint training. Our experiments show that the benefit of joint learning depends on the capabilities of both residual learners. When their standalone performance differs substantially, combining the two provides only limited improvement over the stronger learner. Once noise-space learning is improved to a level comparable to action-space residual learning, joint training can more effectively exploit policy improvements from both spaces.

Our main contributions can be summarized as follows:
\begin{itemize}
\item We identify a mismatch between hard latent-geometric constraints and decoded behavioral quality in noise-space residual learning, and propose \textbf{CNDO} to regulate noise-space adaptation through the soft conservative constraint rather than a fixed latent feasible region.

\item We further introduce \textbf{CANDO}, which jointly optimizes CNDO and action-space residual learning and achieves higher overall sample efficiency than the best single-space method.

\item Through experiments, we reveal the sample-efficiency gap between noise-space and action-space residual learning and further show that joint training provides limited additional benefit when the standalone capabilities of the two residual learners are substantially mismatched.

\end{itemize}

% \item Extensive experiments on the Dual-arm Kitchen Benchmark show that DAG-Plan significantly outperforms other methods. DAG-Plan achieves 52.8\% increase in efficiency over the baseline by directly employing LLMs to generate the single-arm plan. Compared to the baseline, which directly utilizes LLMs for dual-arm plans, DAG-Plan demonstrates 48\% higher success rate. Compared to iterative methods, DAG-Plan improving execution efficiency 84.1\% due to its fewer query time.

%===============================================================================

\section{Related Work}

\subsection{Residual Reinforcement Learning}

Residual reinforcement learning improves an existing controller or policy by preserving its effective behavioral prior and learning only the corrections that are missing from the base policy. Early works on Residual Policy Learning and Residual Reinforcement Learning for Robot Control add trainable residual actions on top of hand-designed or model-based controllers, allowing reinforcement learning to avoid searching over the entire control policy from scratch~\cite{silver2018residual,johannink2019residual}. This paradigm was later extended to contact-rich manipulation with visual observations~\cite{schoettler2020industrial}, shared autonomy~\cite{schaff2020shared}, and settings where the base controller is obtained from demonstrations or motion primitives rather than manually designed~\cite{alakuijala2021rrld,davchev2022residual}. 

\subsection{Noise-Space Policy Steering}

Using latent representations to structure policy optimization predates generative robot policies. PLAS optimizes policies in a VAE-derived latent action space, while OPAL and SPiRL learn temporally extended latent primitives or skill priors from offline data~\cite{zhou2021plas,ajay2021opal,pertsch2021spirl}. LAPO further learns a latent-variable policy through advantage-weighted optimization to generate high-value actions~\cite{chen2022lapo}. With diffusion policies, the Gaussian noise input provides a native latent interface for modifying generated behavior. To the Noise and Back demonstrates that manipulating the diffusion process can effectively incorporate external guidance in shared autonomy~\cite{yoneda2023noise}. More closely related to our setting, DSRL formulates the input noise of a frozen diffusion policy as the action of a latent-space MDP and optimizes it using reinforcement learning~\cite{wagenmaker2025steering}. LPS directly optimizes a latent actor using gradients from an original-action-space critic, backpropagated through a differentiable one-step MeanFlow policy, and constrains the actor output to the Gaussian hypersphere~\cite{im2026lps}. More recently, LP-DS retains samples from the Gaussian prior and learns residual noise perturbations while regulating their magnitude through a Lagrangian trust-region constraint~\cite{simsir2026lagrangian}.

% LPS and CNDO both ultimately evaluate latent changes through decoded actions, but regulate those changes differently. LPS enforces spherical latent geometry and always executes the actor output. CNDO instead learns an unconstrained direction around a Gaussian anchor and uses conservative action-value estimation together with a value-supervised gate, preserving the unmodified pretrained behavior as an explicit fallback. LP-DS, in turn, regulates the anchor-centered perturbation by its magnitude through a Lagrangian trust region. These distinctions motivate comparing geometric restriction, conservative evaluation, and selective execution under a common frozen-policy backbone.

\begin{figure*}[b!]
\vspace{-15pt}
    \centering
    \includegraphics[width=1.0\textwidth]{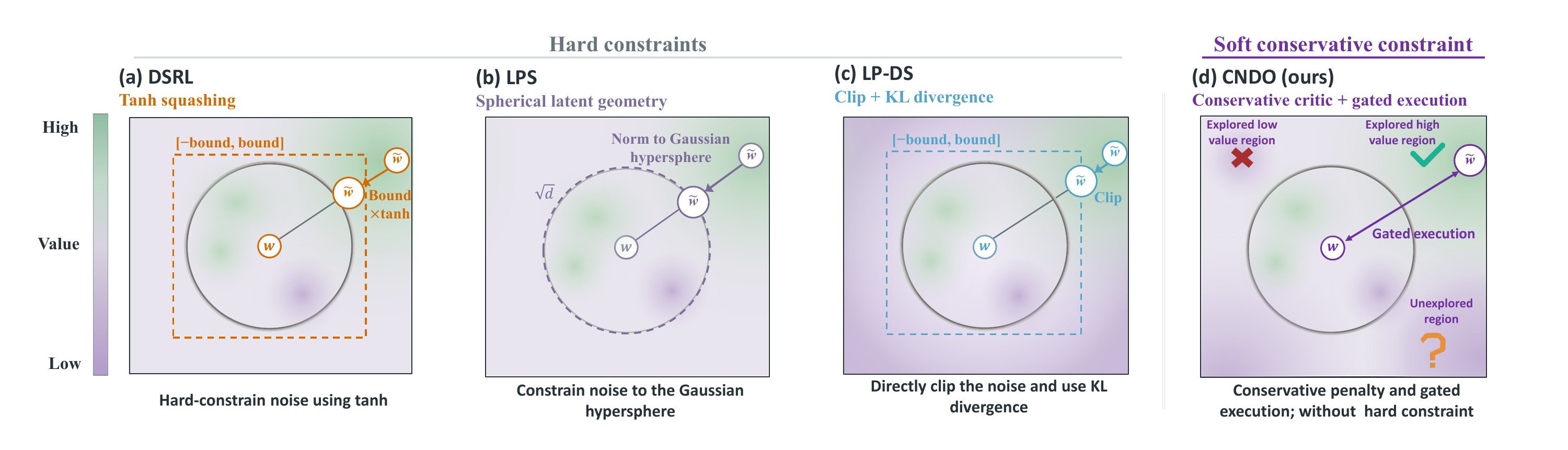}
    \vspace{-29pt}
    \caption{Comparison of noise-space regulation strategies.
    (a) DSRL bounds latent variables with tanh squashing.
    (b) LPS projects latents onto the Gaussian hypersphere.
    (c) LP-DS constrains residual noise with clipping and a trust region.
    (d) CNDO removes fixed geometric constraints and instead uses conservative value estimation and gated execution.}
    \label{fig2}
\end{figure*}

\subsection{Action-Space Residual Reinforcement Learning}

In contrast to noise-space steering before policy decoding, action-space residual learning directly corrects the output action. Recent work has extended the classical residual formulation to high-capacity policies pretrained from offline demonstrations. ResiP learns closed-loop per-step residual corrections over frozen action-chunking policies to compensate for execution errors in precise assembly tasks~\cite{ankile2025refinement}. Policy Decorator introduces a model-agnostic residual adapter for online refinement of large pretrained policies~\cite{yuan2025policydecorator}, while ResFiT applies off-policy RL to learn lightweight per-step corrections over behavior-cloning policies~\cite{ankile2025resfit}. Related approaches also use value functions to coordinate pretrained behaviors with online policy improvement. IBRL treats actions proposed by an imitation policy as candidates for both exploration and temporal-difference bootstrapping~\cite{hu2024ibrl}. DAWN further studies critic cold-start and scale mismatch between base and residual actions as key bottlenecks for value learning in residual RL~\cite{ma2026dawn}. Most closely related to our action-space component, DICE combines residual action optimization with selective behavioral regularization and value-guided multi-candidate action selection, providing a strong sample-efficient action-space baseline~\cite{sun2026from}. USR unifies noise-space steering and action-space residual refinement within a single actor, jointly optimizing latent noise and residual actions, but does not thoroughly investigate the interactions and complementary roles of these two optimization strategies~\cite{zhu2026unified}.

\section{Method}
\label{sec:method}

\begin{figure*}[htbp]
    \centering
    \includegraphics[width=1.0\textwidth]{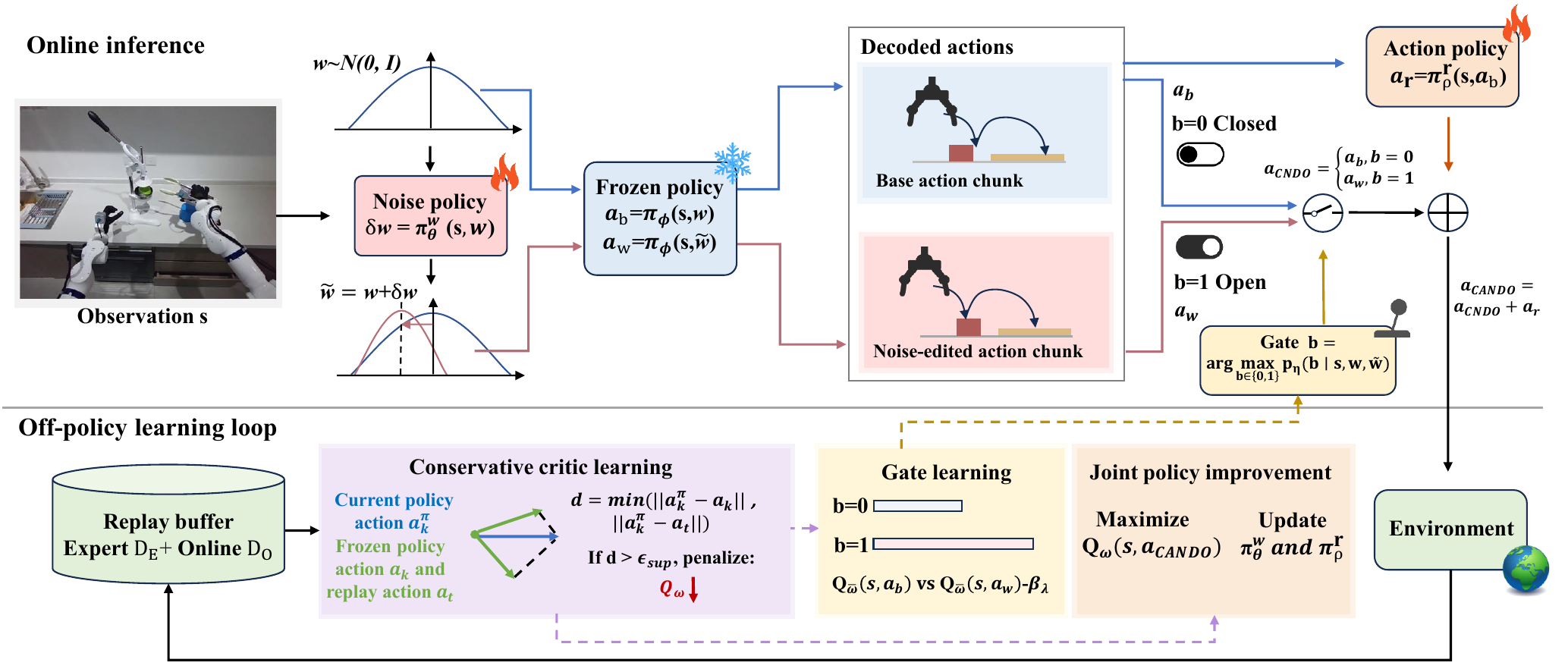}
    \vspace{-15pt}
    \caption{Overview of CANDO.
    During online inference, a noise direction modifies the frozen-policy output, the gate selects whether to use the edit, and an action residual further refines the action.
    During off-policy learning, the critic is trained conservatively, the gate is supervised by the target critic, and both residual branches are jointly optimized.
    The pretrained policy remains frozen throughout.}
    \vspace{-5pt}
\end{figure*}

We develop our method in two variants. First, we introduce \textbf{Conservative Noise Directional Optimization (CNDO)} to improve noise-space reinforcement learning without imposing a fixed feasible region on the edited latent. We then augment CNDO with an action residual to obtain \textbf{Conservative Action-Noise Directional Optimization (CANDO)}, jointly exploiting policy improvement in noise and action spaces. Finally, we summarize the overall framework of CANDO.

\subsection{Conservative Noise Directional Optimization}
\label{sec:cndo}

We consider an action-chunk Markov decision process. Given the observation $s$, a frozen Flow Matching policy generates a base action chunk $a_b$ from Gaussian noise:
\begin{equation}
    w\sim\mathcal{N}(0,I),\qquad
    a_b=\pi_{{\phi}}(s,w),
    \label{eq:base_policy}
\end{equation}
where $\pi_{\phi}$ remains frozen throughout online training. We learn an action-value critic $Q_{\omega}(s,a)$ directly over decoded action chunks, such that noise-space policy improvements are ultimately evaluated by their induced behaviors.

Existing noise-space methods are shown in Fig.~\ref{fig2}. These constraints stabilize policy optimization by explicitly restricting latent exploration. DSRL bounds each latent dimension through a $\tanh$ actor, LPS normalizes latent variables onto the typical hypersphere of a high-dimensional Gaussian, and LP-DS combines residual noise perturbations with explicit bounds and a Lagrangian trust-region constraint~\cite{wagenmaker2025steering,im2026lps,simsir2026lagrangian}. Although these constraints improve stability, latent displacement is only an indirect indicator of decoded behavior quality. Large latent changes may produce valuable behaviors, while small changes do not necessarily imply better actions. Geometric constraints may therefore exclude useful noise-induced policy improvements and limit sample efficiency.

CNDO removes the fixed feasible-region constraint and instead learns a residual direction:
\begin{equation}
\begin{aligned}
    \delta w &= \pi{_{\theta}^w}(s,w), \qquad
    \tilde w=w+\delta w,\\
    a_w &= \pi_{\phi}(s,\tilde w),\qquad
    \delta a_w=a_w-a_b.
\end{aligned}
\label{eq:noise_direction}
\end{equation}
This Gaussian-anchored residual parameterization follows the same general form as LP-DS~\cite{simsir2026lagrangian}. Instead, CNDO focuses on how such directions can be reliably optimized and executed after removing the explicit latent constraint.

Removing the hard constraint enables broader noise-space exploration, but also exposes the critic to decoded actions outside the available behavioral support. Critic extrapolation errors can assign spuriously high values to such actions and drive the direction actor toward increasingly unreliable latent regions. We address this issue with a \textbf{soft conservative constraint} consisting of two complementary components: \textbf{conservative critic} for reliable direction learning and \textbf{gated execution} for selective deployment.

For conservative critic, we maintain an ensemble of $M$ critics and use their minimum whenever a conservative policy or target value is required. The critics are trained from expert and online replay using a multi-sample $n$-step TD objective:
\begin{small}
\begin{equation}
\begin{aligned}
\mathcal{L}_{\mathrm{TD}}
&=
\sum_{j=1}^{M}
\mathbb{E}
\left[
\left(
Q_{\omega_j}(s_t,a_t)-y_t
\right)^2
\right],\\
y_t
&=
R_t^{(n)}
+\gamma^n(1-d_t)
\frac{1}{K_{\mathrm{next}}}
\sum_{k=1}^{K_{\mathrm{next}}}
Q_{\bar{\omega}}(s_{t+n},a'_k).
\end{aligned}
\label{eq:td_loss}
\end{equation}
\end{small}

To further suppress extrapolation errors caused by unrestricted noise edits, we apply conservative regularization only to candidates outside behavioral support:
\begin{small}
\begin{equation}
\begin{aligned}
\mathcal{L}_{Q}
&=
\mathcal{L}_{\mathrm{TD}}
+\alpha_{\mathrm{C}}\mathcal{L}_{\mathrm{C}},\\
\mathcal{L}_{\mathrm{C}}
&=
\sum_{j=1}^{M}
\frac{
\sum_k m_k T
\operatorname{softplus}
\left(
\frac{
Q_{\omega_j}(s_t,a_k^{\pi})
-Q_{\omega_j}(s_t,a_t)
}{T_C}
\right)
}{
\max(\sum_km_k,1)
},\\
m_k
&=
\mathbb{I}
\left[
\min
\left(
\|a_k^{\pi}-a_k\|_2,
\|a_k^{\pi}-a_t\|_2
\right)
>
\epsilon_{\mathrm{sup}}
\right].
\end{aligned}
\label{eq:conservative_q}
\end{equation}
\end{small}
Here, $a_k^{\pi}$ is a detached candidate generated by the current policy, while $a_k$ is the frozen-policy action generated from the same Gaussian anchor. Thus, candidates supported by either the pretrained behavior or replay data remain unpenalized, while unsupported value overestimation is selectively suppressed.

The direction network directly optimizes the complete edited behavior:
\begin{equation}
    \mathcal{L}_{\mathrm{noise}}
    =
    -\mathbb{E}_{s,w}
    \left[
        Q_{\omega}(s,a_w)
    \right]
    +
    \beta_w
    \mathbb{E}_{s,w}
    \left[
        \|\delta w\|_2^2
    \right].
    \label{eq:direction_loss}
\end{equation}
Gradients are propagated through the frozen generative policy to $\pi^{w}_{\theta}$, while the critic and pretrained-policy parameters remain fixed. The second term only provides a weak preference for smaller edits and does not impose a fixed latent feasible region.

Conservative critic reduces unreliable direction optimization, but critic estimates can still be inaccurate, especially early in training. CNDO therefore introduces a chunk-level binary gate $G_\eta$ to regulate the execution of noise edits. Given $(s,w,\tilde w)$, the gate predicts a distribution $p_{\eta}(b\mid s,w,\tilde w)$ over $b\in\{0,1\}$, and the deployed decision is
\begin{equation}
    a_{\mathrm{CNDO}}=\pi_{\mathrm{CNDO}}(s,w)
    =
    a_b+b\delta a_w.
    \label{eq:cndo_policy}
\end{equation}

The gate is trained by distilling the relative values of these two behaviors
from the target critic:
\begin{equation}
    \mathcal{L}_{\mathrm{gate}}
    =
    -\mathbb{E}_{s,w}
    \sum_{b\in\{0,1\}}
    q_{\bar{\omega}}(b\mid s,w,\tilde w)
    \log p_{\eta}(b\mid s,w,\tilde w),
    \label{eq:gate_loss}
\end{equation}
where $q_{\bar{\omega}}$ is the teacher distribution obtained by normalizing and temperature-scaling the target-critic scores $Q_{\bar{\omega}}(s,a_b+b\delta a_w)-\beta_{\lambda}b^{2}$, where \(\beta_{\lambda}\) is the gate-opening penalty factor, such that the noise-edited behavior is favored only when its estimated value improvement is sufficient to offset the gate-opening penalty. Furthermore, we use the more stable target critic to distill the gate, yielding more stable gating decisions than using the online critic either for distillation or for direct action selection.

\subsection{Action-Noise Joint Residual Training}
\label{sec:joint}

CNDO adapts the frozen policy before decoding by editing its input noise. To additionally exploit the post-decoding action space, we introduce an action residual conditioned on the state and action. Specifically, CANDO is defined as
\begin{equation}
\begin{aligned}
    a_{\mathrm{CANDO}} &=\pi_{\mathrm{CANDO}}(s,w,a_b) \\
    &= \pi_{\mathrm{CNDO}}(s,w) + a_r \\
    &=a_b+b\,\delta a_w+a_r,\\
    a_r
    &=
    \pi{_{\rho}^r}
    (
        s,a_b),
    \\
\end{aligned}
\label{eq:cando_policy}
\end{equation}
where $a_r$ denotes the action-space residual.

The two branches are jointly optimized with respect to the final CANDO action:
\begin{small}
\begin{equation}
\begin{aligned}
\mathcal{L}_{\mathrm{joint}}
=&
-\mathbb{E}_{s,w}
\left[
Q_{\omega}(s,a_{CANDO})
\right]\\
&+\beta_w\mathbb{E}_{s,w}
\left[
\|\delta w\|_2^2
\right]+\beta_r\mathbb{E}_{s,w}\left[
\|a_r\|_2^2
\right].
\end{aligned}
\label{eq:joint_loss}
\end{equation}
\end{small}
The first term jointly improves the noise direction and action residual, while the remaining terms weakly regularize their respective magnitudes. CANDO retains CNDO's gate exclusively for noise-space edits, as the gate is designed to regulate noise-space adaptation rather than the action-space residual. The action residual is applied after gating and therefore remains active when $b=0$, in which case the executed action is $a_b+a_r$. We also evaluated alternative conditioning variables for the action residual. 
%Both $\pi_\rho^r(s, a_b)$ and $\pi_\rho^r(s, \delta a_w)$ performed best, while additional noise or action inputs did not yield gains. We retain $\pi_\rho^r(s, a_b)$ for simplicity.

CANDO consequently combines two distinct policy-improvement paths: the noise branch modifies behavior through the frozen generator, whereas the action branch directly refines the decoded action.

\subsection{Overall CANDO Framework}
\label{sec:overall}

We summarize CANDO in Alg.~\ref{alg:cando}. CANDO alternates between off-policy parameter updates and online data collection. During each update, we use RLPD~\cite{ball2023efficient} to sample states and transitions from a mixture of expert and online replay, and sample fresh Gaussian latents to construct the base behavior \(a_b\), the noise-edited behavior \(a_w\), and the action residual \(a_r\). The complete CANDO policy is used to generate next-policy actions and conservative candidates for critic learning. We first update the critics using Eq.~\eqref{eq:conservative_q}, distill the gate from the target critic using Eq.~\eqref{eq:gate_loss}, and jointly optimize the noise-direction and action-residual networks using Eq.~\eqref{eq:joint_loss}. Finally, the target critics are updated by Polyak averaging. The pretrained Flow Matching policy remains frozen throughout training.

% During online inference, CANDO first generates the original and noise-edited behaviors from the current observation, uses the learned gate to determine whether the noise edit should be adopted, and then adds the action residual to obtain the executed action. The resulting transition is stored in the replay buffer for subsequent off-policy updates.

\begin{algorithm}[t]
\caption{Conservative Action-Noise Directional Optimization (CANDO)}
\label{alg:cando}
\begin{algorithmic}[1]

\STATE \textbf{Input:} Expert replay $\mathcal{D}_{E}$,
frozen policy $\pi_{\phi}$,
update steps $U$, and interaction steps $T$

\STATE \textbf{Initialize:} Online replay $\mathcal{D}_{O}$,
direction network $\pi^w_{\theta}$,
action-residual network $\pi^r_{\rho}$,
gate network $G_{\eta}$,
critics $Q_{\omega}$, and target critics
$Q_{\bar{\omega}}$

\WHILE{not converged}

    \LCOMMENT{Off-policy parameter learning}

    \FOR{update step $u=1$ \TO $U$}

        \STATE Sample a batch
        $\mathcal{B}\sim\operatorname{RLPD}
        (\mathcal{D}_{E},\mathcal{D}_{O})$

        \STATE Sample $w\sim\mathcal{N}(0,I)$ and compute
        $a_b=\pi_{\phi}(s,w)$

        \STATE Compute
        $\delta w=\pi{_{\theta}^w}(s,w)$,
        $a_w=\pi_{\phi}(s,w+\delta w)$,
        and $\delta a_w=a_w-a_b$

        \STATE Compute
        $a_r=\pi{_{\rho}^r}(s,a_b)$

        \STATE Infer
        $b=\arg\max_{b\in\{0,1\}}
        p_{\eta}(b\mid s,w,\tilde w)$

        \STATE Construct the deployable CANDO action
        $a=a_b+b\,\delta a_w+a_r$

        \STATE Generate next-policy actions and conservative candidates
        using the complete CANDO policy

        \STATE Update critics $\omega$ using
        Eq.~\eqref{eq:conservative_q}

        \STATE Update gate $\eta$ using
        Eq.~\eqref{eq:gate_loss}

        \STATE Update actors $\theta$ and $\rho$ using
        Eq.~\eqref{eq:joint_loss}

        \STATE Update target critics by Polyak averaging $\bar{\omega} = (1-\tau)\bar{\omega}+\tau\omega$
        
    \ENDFOR

    \LCOMMENT{Online data collection}

    \FOR{interaction step $t=1$ \TO $T$}

    %     \STATE Get current observation $s_t$ and sample
    %     $w_t\sim\mathcal{N}(0,I)$

    %     \STATE Compute
    %     $a_{b,t}=\pi_{\phi}(s_t,w_t)$,
    %     $\delta w_t=\pi{_{\theta}^w}(s_t,w_t)$, and
    %     $\tilde a_{w,t}
    %     =\pi_{\phi}(s_t,w_t+\delta w_t)$

    %     \STATE Compute
    %     $\delta a_{w,t}=\tilde a_{w,t}-a_{b,t}$ and
    %     $a_{r,t}
    %     =\pi{_{\rho}^r}(s_t,a_{b,t})$
        
        \STATE Infer and execute CANDO action $a_t$ 
    %     $b_t=\arg\max_{b\in\{0,1\}}
    %     p_{\eta}(b_t\mid s_t,w_t,w_t+\delta w_t)$

    %     \STATE Execute
    %     $a_t
    %     =a_{b,t}+b_t\,\delta a_{w,t}+a_{r,t}$

        \STATE Add
        $(s_t,a_t,r_t,s_{t+1},d_t)$
        to $\mathcal{D}_{O}$

    \ENDFOR

\ENDWHILE

\end{algorithmic}
\end{algorithm}

\section{Experiments}
\label{sec:experiments}

We design the experiments to answer the following questions:

\begin{itemize}
    \item[$\bullet$] \textbf{Q1:} Can CNDO improve noise-space reinforcement learning to a level comparable to strong action-space residual methods?

    \item[$\bullet$] \textbf{Q2:} How do conservative critic and gated execution contribute to CNDO, and how does CNDO compare with hard constrained directional noise-space methods?
    
    \item[$\bullet$] \textbf{Q3:} How does the standalone capability of noise-space and action-space learners affect the benefit of joint residual training?
    
\end{itemize}

\subsection{Experimental Setup}
\label{sec:experimental_setup}

\textbf{Tasks.}
We use state observations on five manipulation tasks from two benchmarks: \textit{Can}, \textit{Square}, and \textit{Transport} from RoboMimic~\cite{mandlekar2021what}, and \textit{PegInsertionSide} and \textit{StackCube} from ManiSkill~\cite{mu2021maniskill}. 

% \begin{table}[t]
% \footnotesize
% \centering
% \caption{Task protocol. ``Budget'' and ``Eval. interval'' are measured in
% environment transitions.}
% \label{tab:protocol}
% \vspace{-3pt}
% \resizebox{\columnwidth}{!}{
% \begin{tabular}{lrrrr}
% \toprule
% \textbf{Task} & \textbf{Expert traj.} & $H$ & \textbf{Budget} & \textbf{Eval. interval} \\
% \midrule
% Can              & 20  & 4 & 0.20M  & 16k \\
% Square           & 50  & 4 & 0.20M & 16k  \\
% Transport        & 50  & 8 & 1.00M  & 64k  \\
% PegInsertionSide & 200 & 4 & 1.00M  & 64k  \\
% StackCube        & 200 & 4 & 0.50M  & 64k  \\
% \bottomrule
% \end{tabular}}
% \vspace{2pt}
% \vspace{-6pt}
% \end{table}

\begin{figure*}[t]
    \centering
    \includegraphics[width=1.0\textwidth]{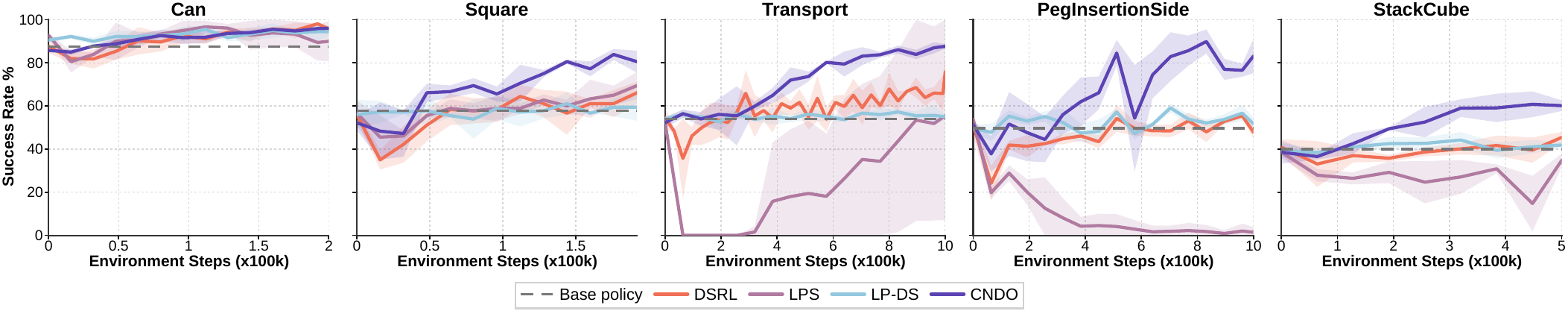}
    \vspace{-20pt}
    \caption{
    Noise-space policy improvement on five tasks. Lines and shaded regions
    show the mean and sample standard deviation over three seeds; the gray
    dashed line is the frozen base policy.
    }
    \label{fig:noise_learning_curves}
\end{figure*}

\begin{figure*}[t]
    \centering
    \includegraphics[width=1.0\textwidth]{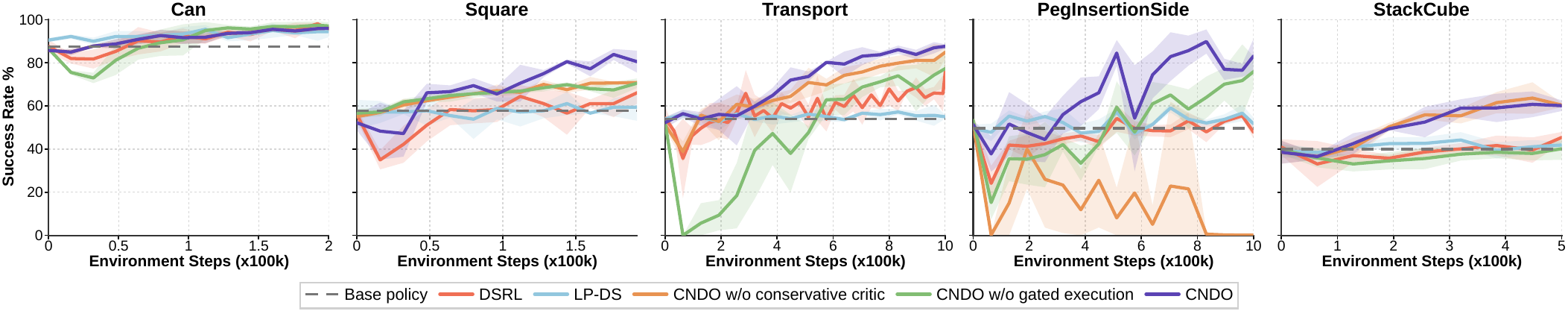}
    \vspace{-20pt}
    \caption{
    CNDO component analysis on five tasks. CNDO ``w/o gated execution'' always executes the
    full latent edit, and ``w/o conservative critic'' sets the behavior-support
    conservative loss weight to zero. Lines and shaded regions denote the mean
    and sample standard deviation over three seeds; the gray dashed line is the
    frozen base policy.
    }
    \label{fig:ablation_learning_curves}
    \vspace{-10pt}
\end{figure*}

\textbf{Online training and evaluation.}
Training uses 16 environments and batch 256 with sparse rewards. Hyperparameters are: updates per collection $U=10$; return steps $n=3$; discount $\gamma=0.99$; critic count $M=10$; next-policy samples $K_{\mathrm{next}}=4$; Polyak rate $\tau=0.01$; conservative regularization weight $\alpha_{\mathrm{C}}=0.1$; conservative regularization temperature $T_C=0.01$; support threshold $\epsilon_{\mathrm{sup}}=1$; noise penalty $\beta_w=10^{-4}$; gate penalty $\beta_{\lambda}=3\times10^{-3}$ and residual action penalty $\beta_r=100$ (RoboMimic) and 3000 (ManiSkill).
We run three seeds (42, 43, and 44) and report the mean and sample standard deviation.

\textbf{Metrics.}
Let $B$ be the max environment steps of each task, and let $\mathrm{SR}(n)\in[0,1]$ denote success rate after $n$ environment transitions. We measure sample efficiency by
\begin{equation}
    \mathrm{NAUC}=\frac{100}{B}\int_0^B \mathrm{SR}(n)\,\mathrm{d}n,
\end{equation}
using linear interpolation between evaluation points and trapezoidal integration. We also report $100\,\mathrm{SR}(B)$ as \textbf{Final SR}. The overall score gives every task equal weight. The frozen behavior-policy result is treated as a constant curve.

\textbf{Baseline algorithms and proposed methods.} We compare the following frozen-policy adaptation methods:

\begin{enumerate}
    \item \textit{Base policy:}
    the pretrained frozen Flow Matching behavior policy without online adaptation.

    \item \textit{DSRL}~\cite{wagenmaker2025steering}:
    a noise-space method that learns latent inputs for the frozen generative policy.

    \item \textit{LPS}~\cite{im2026lps}:
    a latent-policy method that directly propagates action-space value gradients through a differentiable flow decoder while constraining the latent input to the typical Gaussian hypersphere.

    \item \textit{LP-DS}~\cite{simsir2026lagrangian}:
    residual noise steering with an explicit latent bound and a Lagrangian trust-region constraint.

    \item \textit{CNDO}:
    our noise-space directional learner with conservative value estimation and gated execution.

    \item \textit{DICE}~\cite{sun2026from}:
    an action-space residual learner with value-guided candidate selection.

    \item \textit{DSRL+DICE}:
    Combination of DSRL noise learning and DICE action residual learning, following a formulation similar to USR\cite{zhu2026unified}.

    \item \textit{LP-DS+DICE}:
    combination of LP-DS noise learning and DICE action residual learning.

    \item \textit{CANDO}:
    our joint CNDO and action-residual learner.
\end{enumerate}

% one

\subsection{Noise-Space Policy Improvement}
\label{sec:noise_results}

To address \textbf{Q1}, we compare CNDO with existing noise-space methods and use DICE as the action-space reference.
Fig.~\ref{fig:noise_learning_curves} and Tab.~\ref{tab:main_results} show that CNDO consistently improves over the frozen policy and is the strongest noise-space method on average. Relative to DSRL, CNDO raises average NAUC from 57.9 to 69.5 and Final SR from 66.2\% to 81.5\%; relative to LP-DS, the gains are 9.6 NAUC points and 21.0 Final SR points. Relative to the LPS, the gains are 27.8 NAUC points and 31.3 Final SR points. The improvement is particularly large on Transport, PegInsertionSide, and StackCube. Can is the exception in NAUC: its pretrained success rate is already 87.4\%, and LP-DS reaches 93.0 NAUC compared with 91.6 for CNDO and 91.4 for LPS. The LPS port obtains 41.7 average NAUC and 50.2\% average Final SR. The results indicate that spherical latent geometry is not an effective constraint across these five tasks.

CNDO closes the sample-efficiency gap to the action-space reference. The average NAUC of CNDO is 69.5, compared with 68.5 for DICE. But DICE has the higher average Final SR (87.3\% versus 81.5\%), mostly because it reaches 94.1\% on StackCube. Thus, CNDO is the strongest noise-space method and brings noise-space learning to an overall sample-efficiency level comparable to DICE.

\subsection{Ablation Studies of Soft Conservative Constraint}
\label{sec:ablation}

This experiment addresses \textbf{Q2}. We compare full CNDO with different components in Fig.~\ref{fig:ablation_learning_curves} that set the conservative critic weight to zero or force the execution gate to one. We additionally include LP-DS as a geometrically constrained directional-optimization reference; unlike CNDO, LP-DS enforces a fixed latent bound.

Removing conservative critic lowers average NAUC from 69.5 to 58.2. It prevents a severe failure on PegInsertionSide, where Final SR falls from 83.1\% to 0.1\% without it. This supports using the conservative critic to improve stability when the fixed latent feasible region is removed.

Always executing the edit lowers average NAUC to 57.5 and Final SR to 72.2\%. The largest NAUC losses occur on Transport (71.2 to 46.4), PegInsertionSide (65.8 to 49.7), and StackCube (50.9 to 36.6). The gate primarily improves sample efficiency and cross-task robustness, rather than guaranteeing a higher endpoint on every task. Finally, full CNDO exceeds the geometrically constrained LP-DS comparator by 9.6 average NAUC points, consistent with the hypothesis that a fixed latent region can reject useful decoded improvements.

\subsection{Action-Noise Joint Policy Improvement}
\label{sec:joint_results}

\begin{figure*}[t]
    \centering
    \includegraphics[width=1.0\textwidth]{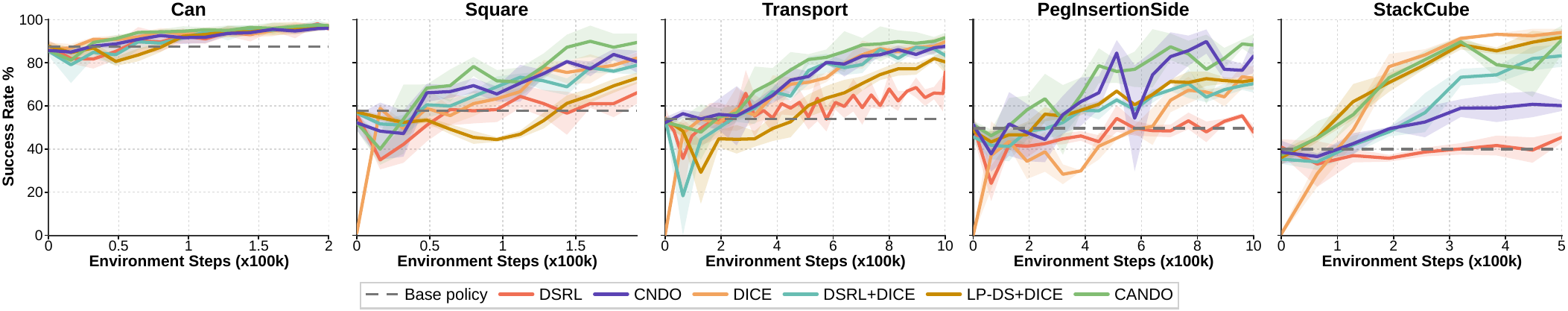}
    \vspace{-15pt}
    \caption{
    Joint noise-action residual learning on five tasks. Lines and shaded
    regions denote the mean and sample standard deviation over three seeds;
    the gray dashed line is the frozen base policy.
    }
    \label{fig:joint_learning}
    \vspace{-2pt}
\end{figure*}

To answer \textbf{Q3}, we compare the standalone and joint versions of the noise-space and action-space learners, with the results shown in Fig. ~\ref{fig:joint_learning}. In particular, we contrast the direct combination DSRL+DICE and LP-DS+DICE with CANDO to determine whether improving the standalone noise learner changes the benefit of joint training.

The standalone DSRL and LP-DS learners trail DICE by 10.6 and 8.6 average NAUC points, respectively. Directly combining either noise-space learner with DICE does not improve the stronger action-space branch: DSRL+DICE obtains 68.0 average NAUC and 82.6\% Final SR, while LP-DS+DICE obtains 67.6 average NAUC and 82.8\% Final SR, both below DICE at 68.5 NAUC and 87.3\% Final SR. At the task level, although joint training provides only marginal or no NAUC improvements on the other tasks, PegInsertionSide deserves particular attention. On this task, DSRL+DICE substantially improves NAUC when DSRL and DICE have comparable standalone performance, and LP-DS+DICE exhibits a similar pattern.

After improving the standalone noise learner, CNDO and DICE are closely matched in average NAUC. Their joint model CANDO reaches 76.1 average NAUC, gains of 6.6 points over CNDO and 7.6 points over DICE, and raises average Final SR to 91.4\%. CANDO ranks first or second on every reported metric except StackCube Final SR, but it does not consistently outperform the single-space methods on every task. This pattern is particularly clear on PegInsertionSide and StackCube. On PegInsertionSide, CNDO substantially outperforms DICE, while CANDO performs similarly to CNDO. The opposite pattern is observed on StackCube, where CANDO performs similarly to DICE. 

These results empirically support the proposed relationship: joint training is most useful here after the two standalone residual learners have comparable capability.

\begin{table*}[t]
\scriptsize
\centering
\caption{
Main comparison results. For each task, \textbf{NAUC} and \textbf{Final SR} are reported in separate columns, in percent, averaged over three seeds. Bold and underlined values denote the best and second-best results, respectively.
}
\label{tab:main_results}
\vspace{-3pt}
\setlength{\tabcolsep}{2.5pt}
\resizebox{\textwidth}{!}{
\begin{tabular}{ll*{6}{cc}}
\toprule
\multirow{2}{*}{\textbf{Method}}
& \multirow{2}{*}{\textbf{Space}}
& \multicolumn{2}{c}{\textbf{Can}}
& \multicolumn{2}{c}{\textbf{Square}}
& \multicolumn{2}{c}{\textbf{Transport}}
& \multicolumn{2}{c}{\textbf{PegInsertionSide}}
& \multicolumn{2}{c}{\textbf{StackCube}}
& \multicolumn{2}{c}{\textbf{Average}} \\
\cmidrule(lr){3-4}\cmidrule(lr){5-6}\cmidrule(lr){7-8}
\cmidrule(lr){9-10}\cmidrule(lr){11-12}\cmidrule(lr){13-14}
& & \textbf{NAUC} & \textbf{Final SR}
& \textbf{NAUC} & \textbf{Final SR}
& \textbf{NAUC} & \textbf{Final SR}
& \textbf{NAUC} & \textbf{Final SR}
& \textbf{NAUC} & \textbf{Final SR}
& \textbf{NAUC} & \textbf{Final SR} \\
\midrule
Base policy
& Base
& / & 87.4
& / & 57.8
& / & 54.0
& / & 49.7
& / & 40.0
& / & 57.8 \\
\midrule
DSRL~\cite{wagenmaker2025steering}
& Noise
& 90.6 & 95.8
& 55.7 & 66.1
& 58.5 & 75.7
& 46.4 & 47.9
& 38.5 & 45.4
& 57.9 & 66.2 \\
LPS~\cite{im2026lps}
& Noise
& 91.4 & 90.0
& 58.0 & 69.4
& 22.5 & 55.4
& 9.1 & 1.4
& 27.3 & 34.8
& 41.7 & 50.2 \\
LP-DS~\cite{simsir2026lagrangian}
& Noise
& \underline{93.0} & 94.4
& 57.5 & 59.4
& 55.2 & 55.0
& 52.5 & 51.8
& 41.2 & 41.9
& 59.9 & 60.5 \\
\rowcolor{gray!15}
\textbf{CNDO}
& Noise
& 91.6 & 96.0
& \underline{68.1} & 80.6
& \underline{71.2} & 87.7
& \underline{65.8} & \underline{83.1}
& 50.9 & 60.1
& \underline{69.5} & 81.5 \\
\midrule
DICE~\cite{sun2026from}
& Action
& 92.9 & \textbf{97.4}
& 63.8 & \underline{82.2}
& 68.4 & \underline{89.8}
& 47.5 & 72.9
& 70.0 & \textbf{94.1}
& 68.5 & \underline{87.3} \\
\midrule
DSRL+DICE
& Joint
& 90.7 & 97.1
& 66.0 & 78.9
& 67.2 & 83.4
& 57.9 & 70.3
& 58.2 & 83.3
& 68.0 & 82.6 \\
LP-DS+DICE
& Joint
& 90.4 & 96.8
& 55.0 & 72.9
& 59.0 & 80.4
& 60.9 & 72.0
& \textbf{72.7} & \underline{91.8}
& 67.6 & 82.8 \\
\rowcolor{gray!15}
\textbf{CANDO}
& Joint
& \textbf{93.2} & \underline{97.2}
& \textbf{72.2} & \textbf{89.4}
& \textbf{74.1} & \textbf{91.7}
& \textbf{70.9} & \textbf{88.2}
& \underline{70.4} & 90.6
& \textbf{76.1} & \textbf{91.4} \\
\bottomrule
\end{tabular}
}
\end{table*}

\begin{figure*}[t!]
    \centering
    \includegraphics[width=1.0\textwidth]{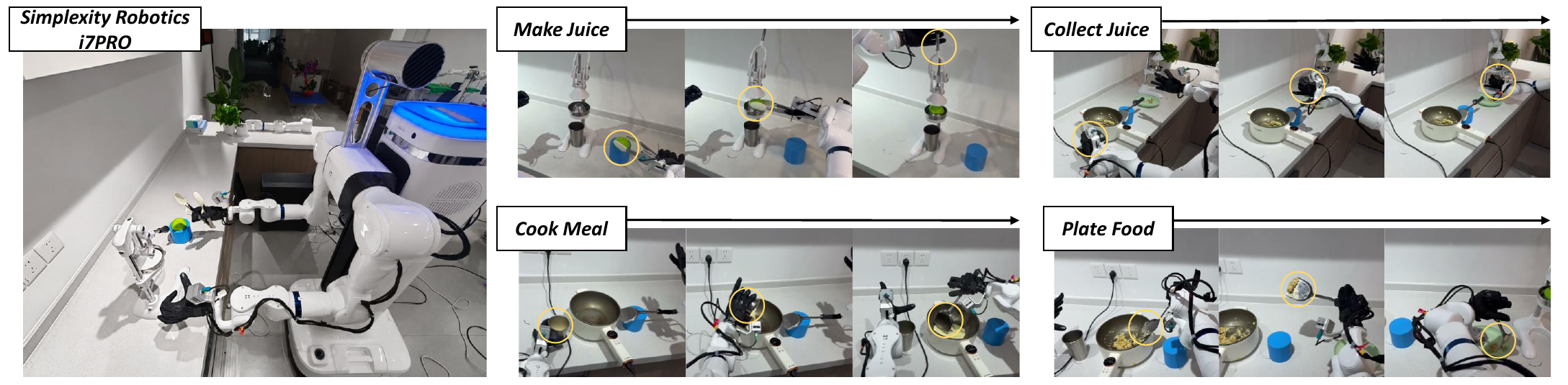}
    \vspace{-15pt}
    \caption{
    Real-world snapshots of our platform and the execution process of tasks.
    }
    \label{real}
    \vspace{-15pt}
\end{figure*}

\subsection{Real-world Experiments}
We validate DSRL, DICE, CNDO, and CANDO on four tasks using the Simplexity Robotics i7PRO dual-arm robot shown in Fig.~\ref{real}, equipped with two Wuji hands, one Gemini 335Lg head camera, and two Gemini 305g wrist cameras. All tasks are performed at fixed positions without position randomization. After each policy rollout an episode, we perform 50 residual reinforcement learning update steps. We roll out 20 episodes for each policy and task. The online training process is provided in the supplementary video and on our project website.

\begin{figure}[htbp]
    \centering
    \includegraphics[width=0.45\textwidth]{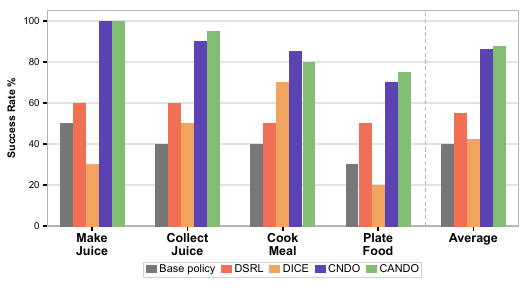}
    \vspace{-20pt}
    \caption{
    Success rates on real-world tasks. We report the success rates measured over 20 evaluation trials for each task.
    }
    \label{real_results}
    \vspace{-10pt}
\end{figure}

As shown in Fig.~\ref{real_results}, CANDO achieves the highest average success rate of 87.5\%, followed closely by CNDO at 86.25\%, while DSRL, DICE, and the frozen base policy achieve 55.0\%, 42.5\%, and 40.0\%, respectively. The success rate of each task in Fig.~\ref{real_results} is measured over 20 independent evaluation trials after residual RL training. The online adaptation process shows that CANDO improves rapidly from the early stage and CNDO continues to improve steadily, whereas DICE exhibits a less consistent trend over prolonged training. Overall, the real-world results resemble the pattern observed on PegInsertionSide: CNDO consistently outperforms DSRL and DICE, while CANDO provides only a small additional gain over CNDO. This result is consistent with our simulation findings: once the noise-space learner itself becomes sufficiently strong, joint training tends to achieve performance comparable to the stronger single-space method, rather than consistently surpassing the best single-space method on every task.

\section{Conclusion}

This work introduced Conservative Noise Directional Optimization (CNDO) and Conservative Action-Noise Directional Optimization (CANDO) for sample-efficient residual RL. CNDO improves noise-space residual learning without imposing a fixed latent feasible region. It optimizes Gaussian-anchored residual directions with conservative value estimation and uses a target-critic supervised gate to selectively execute trusted decoded edits. Across simulation and real-world experiments, CNDO substantially outperforms existing noise-space methods and achieves overall sample efficiency comparable to the strong action-space baseline DICE. Building on this stronger noise-space learner, CANDO jointly optimizes pre-decoding noise directions and post-decoding action residuals at the complete edited action which achieves the highest average sample efficiency. We also figure out that access to both adaptation spaces is not sufficient by itself: joint residual training is most beneficial when the standalone noise-space and action-space learners have comparable capability.

\bibliographystyle{IEEEtran}
\bibliography{main}{}

@article{chi2025diffusion,
  title={Diffusion policy: Visuomotor policy learning via action diffusion},
  author={Chi, Cheng and Xu, Zhenjia and Feng, Siyuan and Cousineau, Eric and Du, Yilun and Burchfiel, Benjamin and Tedrake, Russ and Song, Shuran},
  journal={The International Journal of Robotics Research},
  volume={44},
  number={10-11},
  pages={1684--1704},
  year={2025},
  publisher={Sage Publications Sage UK: London, England}
}

@inproceedings{chisari2025pointflowmatch,
  title     = {Learning Robotic Manipulation Policies from Point Clouds
               with Conditional Flow Matching},
  author    = {Chisari, Eugenio and Heppert, Nick and Argus, Max
               and Welschehold, Tim and Brox, Thomas and Valada, Abhinav},
  booktitle = {Proceedings of the 8th Conference on Robot Learning},
  pages     = {982--993},
  year      = {2025},
  volume    = {270},
  series    = {Proceedings of Machine Learning Research},
  publisher = {PMLR}
}

@inproceedings{ross2011reduction,
  title={A reduction of imitation learning and structured prediction to no-regret online learning},
  author={Ross, St{\'e}phane and Gordon, Geoffrey and Bagnell, Drew},
  booktitle={Proceedings of the fourteenth international conference on artificial intelligence and statistics},
  pages={627--635},
  year={2011},
  organization={JMLR Workshop and Conference Proceedings}
}

@inproceedings{johannink2019residual,
  title     = {Residual Reinforcement Learning for Robot Control},
  author    = {Johannink, Tobias and Bahl, Shikhar and Nair, Ashvin
               and Luo, Jianlan and Kumar, Avinash and Loskyll, Matthias
               and Ojea, Juan Aparicio and Solowjow, Eugen
               and Levine, Sergey},
  booktitle = {2019 International Conference on Robotics and Automation
               (ICRA)},
  pages     = {6023--6029},
  year      = {2019},
  publisher = {IEEE},
  doi       = {10.1109/ICRA.2019.8794127}
}

@inproceedings{ICLR2025_c0749c39,
 author = {Ren, Allen and Lidard, Justin and Ankile, Lars and Simeonov, Anthony and Agrawal, Pulkit and Majumdar, Anirudha and Burchfiel, Benjamin and Dai, Hongkai and Simchowitz, Max},
 booktitle = {International Conference on Learning Representations},
 editor = {Y. Yue and A. Garg and N. Peng and F. Sha and R. Yu},
 pages = {77288--77329},
 title = {Diffusion Policy Policy Optimization},
 volume = {2025},
 year = {2025}
}

@inproceedings{wagenmaker2025steering,
  title     = {Steering Your Diffusion Policy with Latent Space
               Reinforcement Learning},
  author    = {Wagenmaker, Andrew and Zhang, Yunchu
               and Nakamoto, Mitsuhiko and Park, Seohong
               and Yagoub, Waleed and Nagabandi, Anusha
               and Gupta, Abhishek and Levine, Sergey},
  booktitle = {Proceedings of the 9th Conference on Robot Learning},
  pages     = {258--282},
  year      = {2025},
  volume    = {305},
  publisher = {PMLR}
}

@article{simsir2026lagrangian,
  title   = {Lagrangian Perturbation Diffusion Steering:
             Latent Reinforcement Learning for Generative Policies},
  author  = {Simsir, Hikmet and Oguz, Ozgur S.},
  journal = {arXiv preprint arXiv:2606.01151},
  year    = {2026}
}

@inproceedings{
sun2026from,
title={From Prior to Pro: Efficient Skill Mastery via Distribution Contractive {RL} Finetuning},
author={Zhanyi Sun and Shuran Song},
booktitle={Forty-third International Conference on Machine Learning},
year={2026},
pages={1--17}
}

@article{silver2018residual,
  title   = {Residual Policy Learning},
  author  = {Silver, Tom and Allen, Kelsey R. and Tenenbaum, Joshua B.
             and Kaelbling, Leslie Pack},
  journal = {arXiv preprint arXiv:1812.06298},
  year    = {2018}
}

@inproceedings{schoettler2020industrial,
  title     = {Deep Reinforcement Learning for Industrial Insertion Tasks
               with Visual Inputs and Natural Rewards},
  author    = {Schoettler, Gerrit and Nair, Ashvin and Luo, Jianlan
               and Bahl, Shikhar and Ojea, Juan Aparicio
               and Solowjow, Eugen and Levine, Sergey},
  booktitle = {2020 IEEE/RSJ International Conference on Intelligent
               Robots and Systems (IROS)},
  pages     = {5548--5555},
  year      = {2020},
  publisher = {IEEE},
  doi       = {10.1109/IROS45743.2020.9341714}
}

@inproceedings{schaff2020shared,
  title     = {Residual Policy Learning for Shared Autonomy},
  author    = {Schaff, Charles and Walter, Matthew R.},
  booktitle = {Proceedings of Robotics: Science and Systems},
  year      = {2020},
  month     = jul,
  pages     = {1--10},
  doi       = {10.15607/RSS.2020.XVI.072}
}

@article{alakuijala2021rrld,
  title   = {Residual Reinforcement Learning from Demonstrations},
  author  = {Alakuijala, Minttu and Dulac-Arnold, Gabriel
             and Mairal, Julien and Ponce, Jean and Schmid, Cordelia},
  journal = {arXiv preprint arXiv:2106.08050},
  year    = {2021}
}

@article{davchev2022residual,
  title   = {Residual Learning from Demonstration:
             Adapting {DMP}s for Contact-Rich Manipulation},
  author  = {Davchev, Todor and Luck, Kevin Sebastian and Burke, Michael
             and Meier, Franziska and Schaal, Stefan
             and Ramamoorthy, Subramanian},
  journal = {IEEE Robotics and Automation Letters},
  volume  = {7},
  number  = {2},
  pages   = {4488--4495},
  year    = {2022},
  publisher = {IEEE}
}

@inproceedings{zhou2021plas,
  title     = {{PLAS}: Latent Action Space for Offline Reinforcement Learning},
  author    = {Zhou, Wenxuan and Bajracharya, Sujay and Held, David},
  booktitle = {Proceedings of the 2020 Conference on Robot Learning},
  pages     = {1719--1735},
  year      = {2021},
  editor    = {Kober, Jens and Ramos, Fabio and Tomlin, Claire},
  volume    = {155},
  series    = {Proceedings of Machine Learning Research},
  publisher = {PMLR}
}

@inproceedings{ajay2021opal,
  title     = {{OPAL}: Offline Primitive Discovery for Accelerating
               Offline Reinforcement Learning},
  author    = {Ajay, Anurag and Kumar, Aviral and Agrawal, Pulkit
               and Levine, Sergey and Nachum, Ofir},
  booktitle = {International Conference on Learning Representations},
  year      = {2021},
  pages     = {1-23}
}

@inproceedings{pertsch2021spirl,
  title     = {Accelerating Reinforcement Learning with Learned Skill Priors},
  author    = {Pertsch, Karl and Lee, Youngwoon and Lim, Joseph J.},
  booktitle = {Proceedings of the 2020 Conference on Robot Learning},
  pages     = {188--204},
  year      = {2021},
  editor    = {Kober, Jens and Ramos, Fabio and Tomlin, Claire},
  volume    = {155},
  series    = {Proceedings of Machine Learning Research},
  publisher = {PMLR}
}

@inproceedings{chen2022lapo,
  title     = {{LAPO}: Latent-Variable Advantage-Weighted Policy
               Optimization for Offline Reinforcement Learning},
  author    = {Chen, Xi and Ghadirzadeh, Ali and Yu, Tianhe
               and Wang, Jianhao and Gao, Alex Yuan and Li, Wenzhe
               and Bin, Liang and Finn, Chelsea and Zhang, Chongjie},
  booktitle = {Advances in Neural Information Processing Systems},
  volume    = {35},
  pages     = {36902--36913},
  year      = {2022}
}

@inproceedings{yoneda2023noise,
  title     = {To the Noise and Back: Diffusion for Shared Autonomy},
  author    = {Yoneda, Takuma and Sun, Luzhe and Yang, Ge
               and Stadie, Bradly C. and Walter, Matthew R.},
  booktitle = {Proceedings of Robotics: Science and Systems},
  year      = {2023},
  pages     = {1--18}
}

@inproceedings{im2026lps,
  title     = {Latent Policy Steering through One-Step Flow Policies},
  author    = {Im, Hokyun and Kolobov, Andrey and Fu, Jianlong
               and Lee, Youngwoon},
  booktitle = {Proceedings of Robotics: Science and Systems},
  year      = {2026},
  pages     = {1--14}
}

@inproceedings{ankile2025refinement,
  title     = {From Imitation to Refinement:
               Residual {RL} for Precise Assembly},
  author    = {Ankile, Lars and Simeonov, Anthony and Shenfeld, Idan
               and Torne, Marcel and Agrawal, Pulkit},
  booktitle = {2025 IEEE International Conference on Robotics
               and Automation (ICRA)},
  pages     = {1--8},
  year      = {2025},
  publisher = {IEEE}
}

@inproceedings{yuan2025policydecorator,
  title     = {Policy Decorator: Model-Agnostic Online Refinement
               for Large Policy Model},
  author    = {Yuan, Xiu and Mu, Tongzhou and Tao, Stone
               and Fang, Yunhao and Zhang, Mengke and Su, Hao},
  booktitle = {International Conference on Learning Representations},
  year      = {2025},
  pages     = {1-36}
}

@article{ankile2025resfit,
  title   = {Residual Off-Policy {RL} for Finetuning
             Behavior Cloning Policies},
  author  = {Ankile, Lars and Jiang, Zhenyu and Duan, Rocky
             and Shi, Guanya and Abbeel, Pieter and Nagabandi, Anusha},
  journal = {arXiv preprint arXiv:2509.19301},
  year    = {2025}
}

@inproceedings{hu2024ibrl,
  title     = {Imitation Bootstrapped Reinforcement Learning},
  author    = {Hu, Hengyuan and Mirchandani, Suvir and Sadigh, Dorsa},
  booktitle = {Proceedings of Robotics: Science and Systems},
  year      = {2024},
  pages     = {1-18}
}

@article{ma2026dawn,
  title   = {What Makes Value Learning Efficient
             in Residual Reinforcement Learning?},
  author  = {Ma, Guozheng and Li, Lu and Wang, Haoyu and Liu, Zixuan
             and Bacon, Pierre-Luc and Tao, Dacheng},
  journal = {arXiv preprint arXiv:2602.10539},
  year    = {2026}
}

@inproceedings{
mandlekar2021what,
title={What Matters in Learning from Offline Human Demonstrations for Robot Manipulation},
author={Ajay Mandlekar and Danfei Xu and Josiah Wong and Soroush Nasiriany and Chen Wang and Rohun Kulkarni and Li Fei-Fei and Silvio Savarese and Yuke Zhu and Roberto Mart{\'\i}n-Mart{\'\i}n},
booktitle={5th Annual Conference on Robot Learning },
pages     = {1--13},
year={2021},
publisher = {PMLR}
}

@inproceedings{mu2021maniskill,
  title     = {ManiSkill: Generalizable Manipulation Skill Benchmark with Large-Scale Demonstrations},
  author    = {Mu, Tongzhou and Ling, Zhan and Xiang, Fanbo and Yang, Derek Cathera
               and Li, Xuanlin and Tao, Stone and Huang, Zhiao
               and Jia, Zhiwei and Su, Hao},
  booktitle = {Proceedings of the Neural Information Processing Systems Track on Datasets and Benchmarks},
  volume    = {1},
  pages     = {1--14},
  year      = {2021}
}

@inproceedings{ball2023efficient,
  title={Efficient online reinforcement learning with offline data},
  author={Ball, Philip J and Smith, Laura and Kostrikov, Ilya and Levine, Sergey},
  booktitle={International Conference on Machine Learning},
  pages={1577--1594},
  year={2023},
  organization={PMLR}
}

@inproceedings{zhu2026unified,
  title={Unified Latent Steering and Residual Refinement for Online Improvement of Diffusion Policy Models},
  author={Zhu, Zhengbang and Li, Ziyan and Yuan, Xiu and Zhang, Hanbo and Liu, Yuxiao and Zhang, Chongjie and Yu, Yong and Zhang, Weinan and Liu, Minghuan},
  booktitle={ICML 2026 Workshop on Decision-Making from Offline Datasets to Online Adaptation},
  year={2026},
}
 
\end{document}